\documentclass[conference]{IEEEtran}
\IEEEoverridecommandlockouts
\usepackage{cite}
\usepackage{amsmath,amssymb,amsfonts}
\usepackage{algorithmic}
\usepackage{graphicx}
\usepackage{textcomp}
\usepackage{xcolor}
\def\BibTeX{{\rm B\kern-.05em{\sc i\kern-.025em b}\kern-.08em
    T\kern-.1667em\lower.7ex\hbox{E}\kern-.125emX}}

\usepackage[ruled,vlined]{algorithm2e}
\usepackage{bm}
\usepackage{amsmath}

\usepackage{comment}
\usepackage{todonotes}
\usepackage{pifont}

\usepackage{tikz}
\usepackage{amsmath}

\usepackage{filecontents}

\usepackage{subfigure}

\usepackage{booktabs}
\usepackage{graphicx}
\usepackage{tabularx}

\definecolor{darkblue}{rgb}{0.18,0.14,0.54}
\definecolor{sblack}{rgb}{0,0,0}

\begin{document}
\title{STAG: Enabling Low Latency and Low \underline{Sta}leness of \underline{G}NN-based Services with Dynamic Graphs\\
}

\author{
\IEEEauthorblockN{Jiawen Wang\textsuperscript{1}, Quan Chen\textsuperscript{1}, Deze Zeng\textsuperscript{2}, Zhuo Song\textsuperscript{3}, Chen Chen\textsuperscript{1} and Minyi Guo\textsuperscript{1}}
\IEEEauthorblockA{\textsuperscript{1} \textit{Dept. of Computer Science and Engineering, Shanghai Jiao Tong University, Shanghai, China}\\
\textsuperscript{2} \textit{School of Computer Science, China University of Geosciences, Wuhan, China}\\
\textsuperscript{3} \textit{Alibaba Cloud, Hangzhou, China}\\
\{wangjiawen0606, chen-quan\}@sjtu.edu.cn}, deze@cug.edu.cn, songzhuo.sz@alibaba-inc.com, \{chen-chen, guo-my\}@sjtu.edu.cn
}

\maketitle

\begin{abstract}
Many emerging user-facing services adopt Graph Neural Networks (GNNs) to improve serving accuracy. 
When the graph used by a GNN model changes, representations (embedding) of nodes in the graph should be updated accordingly.
However, the node representation update is too slow, resulting in either long response latency of user queries (the inference is performed after the update completes) or high staleness problem (the inference is performed based on stale data).

Our in-depth analysis shows that the slow update is mainly due to neighbor explosion problem in graphs and duplicated computation. 
Based on such findings, we propose \underline{{\bf STAG}}, a GNN serving framework that enables low latency and low staleness of GNN-based services. 
It comprises a \textbf{\textit{ collaborative serving mechanism}} and an \textbf{\textit{ additivity-based incremental propagation strategy}}.
With the collaborative serving mechanism,
only part of node representations are updated during the update phase,  
and the final representations are calculated in the inference phase.
It alleviates the neighbor explosion problem.
The additivity-based incremental propagation strategy reuses intermediate data during the update phase, eliminating duplicated computation problem. 
Experimental results show that STAG accelerates the update phase by 1.3x$ \sim $90.1x, and 
greatly reduces staleness time with a slight increase in response latency.
\end{abstract}


\section{Introduction}
\label{sec:intro}

User-facing services (e.g., e-commerce~\cite{apan}, social network~\cite{social3}, recommendation system ~\cite{recommend4}) now widely adopt Graph Neural Networks (GNNs) as a component to learn knowledge from large-scale graphs.
Representation of a node shows the abstracted features of the node in the GNN model.
When a user query is received, serving system obtains corresponding nodes' representations and generates inference results.

\begin{figure}[t]
    \centering \includegraphics[width=.9\columnwidth]{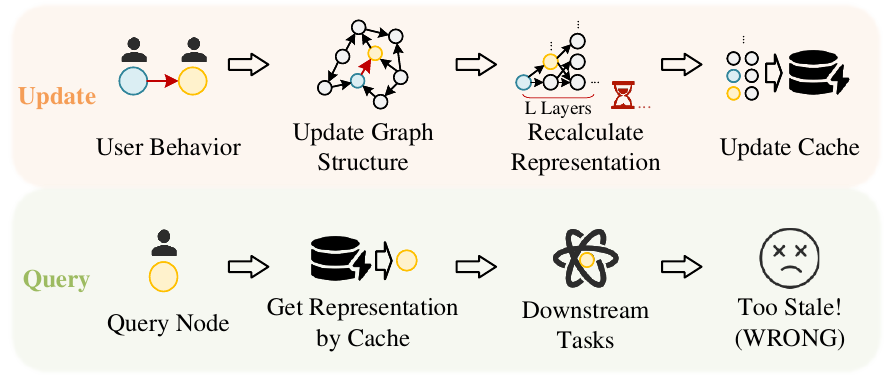}
    \caption{\label{fig:GetLatestEmbedding} Serving a GNN service based on dynamic graphs.}
    \vspace{-3.5mm}
\end{figure}

Many real productions (e.g., recommendation systems~\cite{survey_recommend} and fraud detection systems~\cite{fraud_detection}) rely on dynamic graphs~\cite{survey_dynamic, ebay, apan}, in which new nodes or edges may be added to or deleted from the graph, and the attributes of a node may change as well. 
Figure \ref{fig:GetLatestEmbedding} shows a typical serving system that hosts a GNN-based user-facing service with a dynamic graph~\cite{cb}.
In general, system receives two kinds of requests, i.e., \textit{update} and \textit{query}.
Update requests are triggered by graph change events and Queries are submitted by end users. 
They may happen simultaneously.
When a user query is received, serving system returns previously saved/cached representations of the corresponding nodes directly with low latency. 
When an update request is received, 
backend graph maintainer updates graph structure and cached representations of affected nodes. The affected nodes are determined by the graph change event~\cite{cb}.
For instance, if the raw feature of node $v$ in an $L$-layer GNN model with is modified, all the nodes that can be reached within $L$-hop from $v$ will be affected and all their representations shall be updated.
This is because each layer in GNNs aggregates information from neighbors of a node. Representation of a node reflects the information of its neighbors within $L$-hop in an $L$-layer GNN.

It is time-consuming to update node representations, especially when the connectivity (fan-in/fan-out edges) is high and GNN model has many layers.
For instance, with graph in Reddit dataset~\cite{reddit} and a 3-layer GCN model, more than 10 seconds are required to calculate node representations under minor changes of graph. 
In this case, user queries may get wrong results due to the stale node representations that do not reflect the latest changes in graph. 
This problem refers to {\it high staleness problem}.
Due to the high cost of updating node representations in real-time, some older industrial schemes accumulate graph changes, and periodically (e.g., in days) updates the node representations at once ~\cite{recommend1, fraud_detection2}. They suffer from worse high staleness problem.

The high staleness problem could be fatal~\cite{apan,recom_staleness}. For instance, in an e-commerce fraud detection system, fraudsters always carry out a series of fraud transactions and withdraw the money quickly~\cite{apan}. Ideally, fraud detection system should recognize fraudsters by the first few fraud transactions and ban them as soon as possible, preferably within seconds. However, due to high staleness, the updated representation that contains information of fraud transactions may not be available even after fraudsters escape. 

Some prior work~\cite{aligraph} tried to solve the high staleness problem by letting online query to perform inference directly using GNN model, without relying on cached representations. 
However, this method leads to long response latency due to the slow inference.
For instance, 5 seconds are needed to perform a GNN inference for one single node on Reddit~\cite{reddit} with a 3-layer GCN model.
Moreover, repeated requests for the same node or adjacent neighboring nodes also bring duplicated computation. 
The high staleness becomes high latency in this case, and it refers to the \textit{high latency problem}.


A better solution is to speed up update/inference directly. Our investigation shows that there are two issues that result in the slow update/inference: 
First, the number of to-be-updated node representations increase explosively with the number of layers in GNN model and the connectivity of graph.
For instance, if each node has 500 neighbors in a graph (in Reddit~\cite{reddit}) and a 2-layer GNN model is used, the representations of approximately $500^2=250000$ nodes should be updated.
Secondly, existing update method is not efficient since it involves unaffected nodes, which introduces unnecessarily duplicated computation.

Based on the above findings, we propose a novel GNN serving framework {\bf STAG}. It achieves both low staleness and low latency
through a {\it collaborative serving mechanism} ({\bf CSM}) and an {\it additivity-based incremental propagation strategy} ({\bf AIP}). 
With CSM, when a node's raw feature is updated, backend graph maintainer only updates and saves $M$-layer representations of nodes that are reached in $M$-hop ($0 \le M \le L$). 
When a query is received, a frontend serving handler calculates the remaining $L-M$-layer representations of the corresponding nodes based on the saved representations. 
Together with a proposed policy that tuning an appropriate $M$, CSM balances the workload between graph maintainer and serving handler. CSM resolves the first issue that results in slow update/inference.
Moreover, we find that the aggregation operations in GNN models usually exhibit additivity property. 
Based on this property, AIP reuses intermediate data during the update phase, thus eliminating the duplicated computation. AIP resolves the second issue that results in slow update/inference.

The contributions of this paper are as follows.
\begin{itemize}
    \item The comprehensive discussion of low staleness and low latency dilemma. The discussion motivates us to design a GNN serving system to minimize the staleness while ensuring low service latency.

    \item The design of collaborative serving mechanism (CSM) and corresponding policies. CSM alleviates neighbor explosion problem with the collaboration of graph maintainer and forward inference. Different policies can be designed based on the mechanism to achieve different optimization purposes (e.g., ensuring the given staleness or service latency).

    \item The finding of additivity property of the intermediate data in GNN models. This finding enables the design of our additivity-based incremental propagation strategy that eliminates duplicated computation. 

\end{itemize}
Experimental results show that STAG enables low staleness and low latency simultaneously, and can support 2.7x$\sim$27x workloads compared with existing approaches.

\section{Related works}\label{sec:related work}

Graph Neural Networks (GNNs) have been widely-used for representation learning on graph-structured data \cite{survey_comprehensive, survey_accelerator, survey_dynamic}. A large number of prior works~\cite{p3, featgraph, aligraph} have been done for improving the efficiency of GNN training, while the studies focusing on GNN inference are much less than those focusing on training.

Several studies have attempted to optimize computation procedure of GNN or design hardware accelerators, such as HAG~\cite{hag}, PCGCN~\cite{pcgcn}, and GRIP~\cite{grip}.
\textcolor{black}{However, these studies rarely considered the problems that the dynamism of the graph brings to GNN services, making them difficult to be applied on dynamic graphs.}

CB~\cite{cb} and Aligraph~\cite{aligraph} are capable to adapt the graph dynamics and perform dynamic inference on common GNN models, similar to our work.
CB~\cite{cb} determined the influenced nodes precisely by propagating the information of incoming graph dynamics, and updated the influenced nodes without performing inference on entire graph. Also, they develop a cache system to reuse the hidden representations in previous inference.
Aligraph~\cite{aligraph} proposed a subsystem of Dynamic Graph Service to enable the online GNN service on dynamic graphs. Instead of maintaining the cache of node representations, Aligraph obtains the representations by dynamic sampling and neighbor aggregation. 
\textcolor{black}{However, they either suffer from high staleness problem or high latency problem (discussed in detail in Section ~\ref{sec:moti}).
}

\section{Background and Motivation}
\label{sec:moti}


\subsection{Serving GNN on Dynamic Graphs}
\label{sec:background}

\begin{figure}
\centerline{\includegraphics[width=\columnwidth]{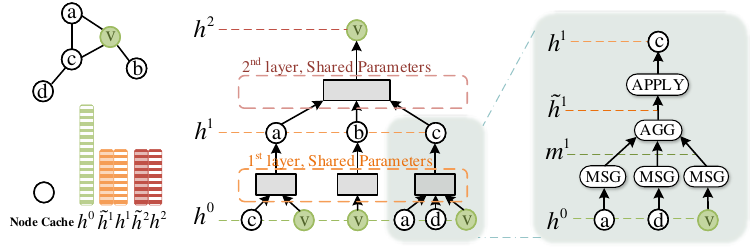}}
\caption{An example of GNN inference.}
\label{fig:gnninf}
\end{figure}

Let $\mathcal{G}=(\mathrm{V},\mathrm{E})$ represents a graph, where $\mathrm{V}$ and $\mathrm{E}$ represent the sets of nodes and edges, respectively.
$(u,v)\in E$ denotes the edge from  node $u$ to  node $v$.
The raw feature of node $v$, denoted by $h^{0}(v)$, is a high-dimensional vector that represents the attributes of the entity corresponding to the node. For an $L$-layer GNN model, it recursively aggregates $L$-hop neighborhood for each node to generate its representation. The middle of Figure \ref{fig:gnninf} shows an example that calculates representation of node $v$ using a 2-layer GNN.

As shown in the right part of Figure~\ref{fig:gnninf}, a node's representation in a layer is generated in three stages, {\it MSG} (message), {\it AGG} (aggregate), and {\it APPLY}. 
{\it MSG} is an edge-wise operation, it generates a message on each edge using the source and target node features of the edge.  
{\it AGG}  aggregates the messages on each in-edge to generate a neighborhood representation.
{\it APPLY} performs dimension reduction using neural networks to generate hidden representation in that layer.

The definitions of the three operations are shown in Equation~\eqref{eq:msg_passing}, where $\mathrm{IN}(v)$ denotes the in-neighbor node set of $v$;
$m^{l}(u,v)$ is the message on edge $(u,v)$ at layer $l$; $\widetilde{h}^{l}(v)$ denotes the aggregated neighbor information of node $v$ at layer $l$; $h^l(v)$ is the $l$-th layer representation of node $v$, especially, $h^0(v)$ refers to the raw feature of node. 
The representation of the last layer $h^L(v)$ is usually called node representation (or node embedding), while others are called hidden representations. 
\begin{equation}
\label{eq:msg_passing}
\begin{aligned}
 & m^{l+1}(u,v) = \mathrm{MSG}(h^l(u),h^l(v)), (u,v)\in E ; \\
 & \widetilde{h}^{l+1}(v)=\mathrm{AGG}(\{m^{l+1}(u,v) | u \in \mathrm{IN}(v)\}) ; \\
 & h^{l+1}(v) = \mathrm{APPLY}(\widetilde{h}^{l+1}(v),h^l(v)) 
\end{aligned}
\setlength{\abovedisplayskip}{--3mm}
\setlength{\belowdisplayskip}{--3mm}
\end{equation}

Whenever a graph is updated (e.g., adding a new node, modifying node feature, etc.), the representations of some nodes in the graph become out-of-date~\cite{cb}. 
For instance, if a node $v$'s feature is modified and a 2-layer GNN model is used,  representations of all nodes that can be reached within 2-hop from node $v$ should be updated.

\subsection{Existing Representation Update Solutions}
\label{subsec:weakness}

There generally exists two types of approaches, forward inference-based ({\it inf-based}) approach~\cite{aligraph} and backend update-based ({\it upd-based}) approach~\cite{cb} for updating the representations on  dynamic graphs. 
Figure~\ref{fig:example_aggr_prop} shows an example of updating a node's representation in a 2-layer GNN model.

With inf-based approach, backend graph maintainer only updates graph's structure when receiving update requests. 
When frontend serving handler receives a query, 
it samples a $2$-layer subgraph corresponding to the query, and 
uses GNN model to perform inference on the subgraph to obtain the queried representations.
This approach ensures that the user query is always done based on the non-stale data. 
Aligraph~\cite{aligraph} adopts the inf-based approach.

With upd-based approach, when receiving update requests, graph maintainer are responsible for updating representations of all nodes that can be reached within $2$-hop
(Figure~\ref{fig:example_aggr_prop}(b)).
The representations are then cached for user queries. Frontend serving handler directly obtains the required node representations and returns inference result. 
Obviously, this approach ensures low response latency of node representation queries.
CB~\cite{cb} adopts the upd-based approach.

\begin{figure}
    \centering
    \subfigure[Inf-based Approach]
    {
    \includegraphics[width=0.38\linewidth]{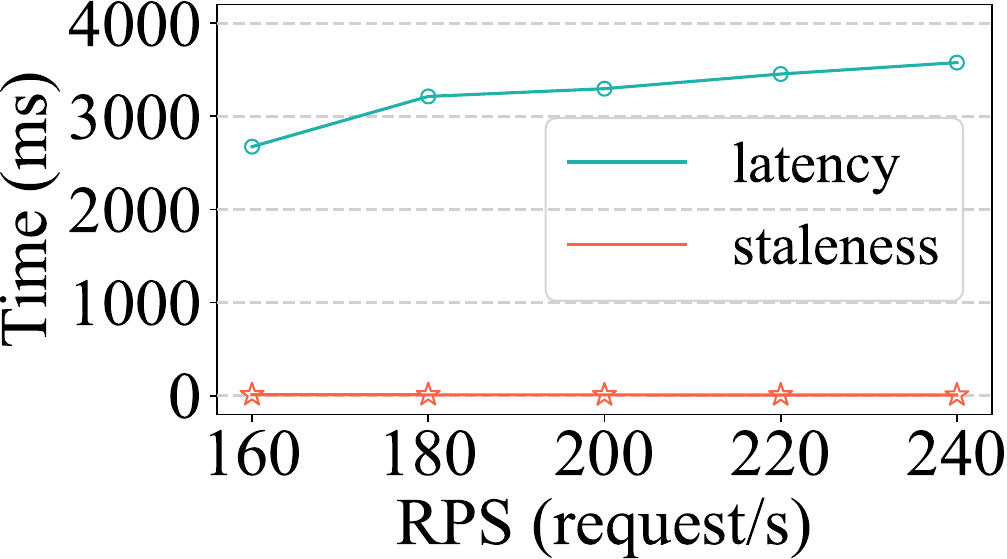}
    }
    \hspace{3mm}
    \subfigure[Upd-based Approach]{
    \includegraphics[width=0.38\linewidth]{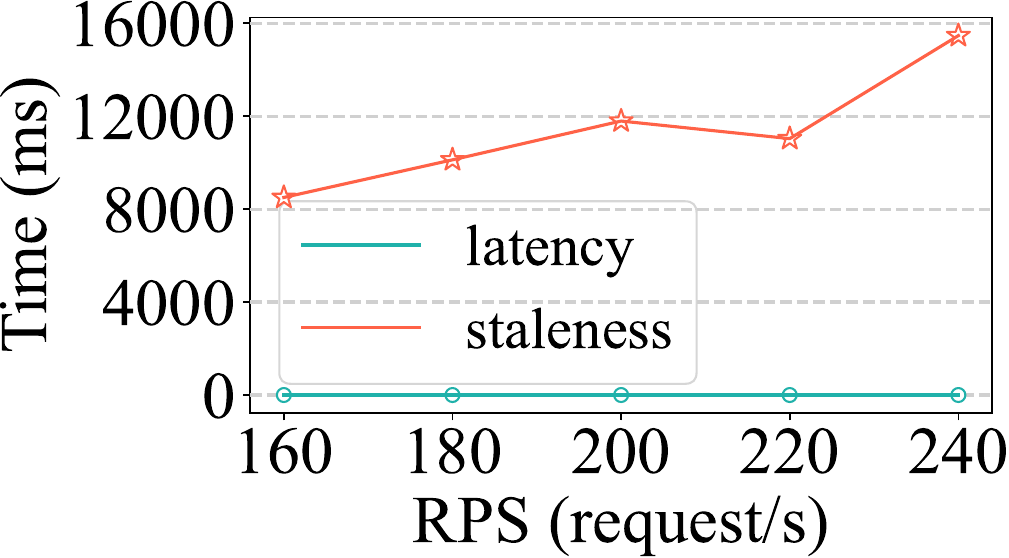}
    }
    \caption{Staleness time and response latency of user queries with the two approaches at different loads. 
    }
    \label{fig:moti_lat_sta}
\end{figure}

We measure both response latency and staleness time of user queries on the updated nodes with Aligraph and CB in Figure \ref{fig:moti_lat_sta}.
The {\it staleness time} of a query is defined as the time between the timestamp that query returned and the timestamp that the node representations are affected.
The larger the staleness time is, the more possible that query gets wrong results.
With Reddit graph~\cite{reddit} and a 2-layer GCN model, the response latency is more than 3 seconds with Aligraph~\cite{aligraph}, and the staleness time is more than 8 seconds with CB~\cite{cb}.

The two approaches result in either long latency or long staleness time.
The time will be much longer in production, where graph is denser and GNN model has more layers.

\subsection{Weakness of Existing Solutions}

\subsubsection{Neighbor Explosion Problem}

\begin{table}
\setlength{\belowcaptionskip}{5.pt}
\centering
\caption{The size of $L$-hop neighborhood (i.e., the number of nodes) in different datasets. }
\label{tab:num_nodes}
\small
\begin{tabular}{@{}ccccccc@{}}
\toprule
\textbf{Hops} & \textbf{Reddit} & \textbf{PPA}   & \textbf{Products} & \textbf{Citation2} & \textbf{dReddit} & \textbf{dWiki} \\ \midrule
\textbf{0}      & 1      & 1     & 1        & 1        & 1       & 1 \\
\textbf{1}      & 483    & 77    & 51       & 21       & 126     & 32 \\
\textbf{2}      & 48864  & 4099  & 3735     & 1456     & 10386 & 8181 \\ \bottomrule
\end{tabular}
\end{table}

As shown in Figure~\ref{fig:example_aggr_prop}, when a node is updated, the number of involved nodes increases exponentially with the number of layer in GNN model and the connectivity of graph. It is time-consuming to update representations of all involved nodes.

As examples, Table~\ref{tab:num_nodes} shows average number of nodes that can be reached in 1/2-hop from a node, with six widely-used graph datasets. 
Assume we update a node in the graph {\it Reddit} and a 2-layer GNN model is used.
The representations of 48,864 nodes should be updated if the upd-based approach is adopted. 
Similarly, a node's representation can be calculated by aggregating information from its 48,864 neighbor nodes within 2-hop using the inf-based method.

STAG alleviates neighbor explosion problem by balancing the workload of forward inference and backend update using {\it CSM} (see Section \ref{sec:collaborative inference}).

\subsubsection{Duplicated Computation in Update phase}
\label{subsubsec:redundant}

According to Equation~\eqref{eq:msg_passing}, if a node's representation becomes stale, existing update approaches~\cite{cb} need to re-compute 
all information of its neighbors (i.e., $m^{l+1}(u,v)$) 
and aggregate them together to update its representation. 

We find that there lies a large amount of duplicated computation, because the information of most neighbors is very likely invariant since last computation. However, these invariant neighbor messages are recalculated and reaggregated duplicatedly in existing update method. We will explain the duplicated computation in detail in Section~\ref{sec:aip}.

STAG eliminates above duplicated computation by updating representation incrementally with the observed addivity property in {\it AGG} phase
using {\it AIP} strategy (see Section \ref{sec:aip}).

\section{Overview of STAG}
\label{sec:design}
\begin{figure}
\centerline{\includegraphics[width=\columnwidth]{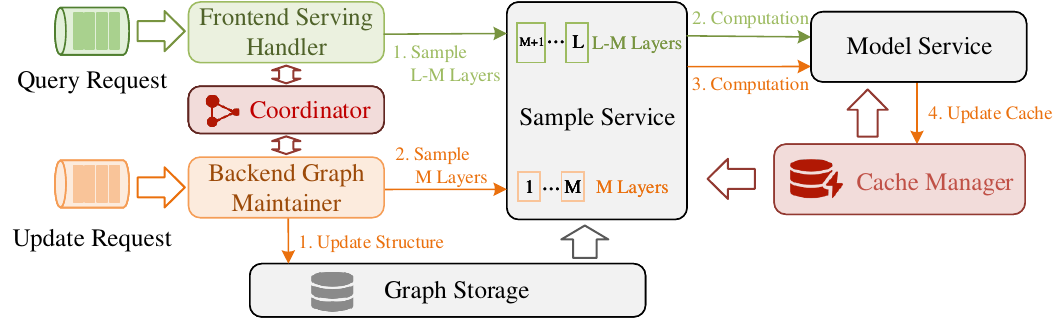}}
\caption{The overall architecture of STAG.} \label{arch}
\end{figure}

Figure~\ref{arch} shows the overall architecture of STAG. 
It has a backend graph maintainer that receives graph update requests, a frontend serving handler that responds to user queries, a {\it coordinator} that balances the workloads of graph maintainer and serving handler and a cache manager stores intermediate node representations calculated by the graph maintainer.

STAG hosts a GNN service based on a dynamic graph and an $L$-layer GNN model in the following steps. First, as the core controller of {\bf CSM} (Section \ref{sec:collaborative inference}), coordinator dynamically determines an appropriate $M$ ($0\le M\le L$) to balance the load of graph maintainer and serving handler, based on the characteristics of graph, graph update frequency, query frequency etc.

When a graph update request is received, graph maintainer updates graph structure (orange lines in Figure~\ref{arch}). Suppose a node is updated, maintainer obtains all  neighbors that can be reached within $M$-hop from the updated node, recalculates neighbors' representations using a slightly modified GNN model realizing {\bf AIP} strategy (Section \ref{sec:aip}), and stores the updated representations in cache. 

When an user query is received, serving handler calculates the final node representations and generates inference result by calling the GNN model (the green lines in Figure~\ref{arch}).
Specifically, serving handler reads intermediate result of $M$-layer from the cache manager as input and finishes up the remaining $L-M$-layer computation.


Note that caching of intermediate results does not incur large memory overhead, as the original representation of nodes is usually of high dimensions, often in hundreds or even thousands of dimensions~\cite{recommend1}, while the dimensions of intermediate results are only around a few dozen. We evaluate the memory overhead for caching in Section~\ref{subsec:scale}. 

\section{Collaborative Serving Mechanism}
\label{sec:collaborative inference}

In this section, we detail the collaborative serving mechanism (CSM), and the way to find the appropriate $M$, i.e., the number of layers done by the graph maintainer during update.

\subsection{Formalizing The Collaborative Serving}
\label{sec:cserving}
\begin{figure}
    \centering
    \subfigure[Inf-based]
    {
    \includegraphics[width=0.27\linewidth]{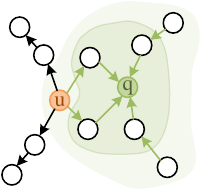}
    }
    \subfigure[Upd-based]{
    \includegraphics[width=0.27\linewidth]{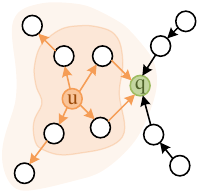}
    }
    \subfigure[STAG with $M=1$]{
    \includegraphics[width=0.27\linewidth]{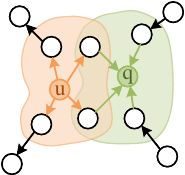}
    }
    \caption{Forward inference (green) and backend update (orange) on a 2-layer GNN model (an update to node $u$ and a query to node $q$).}
    \label{fig:example_aggr_prop}
\end{figure}

Figure \ref{fig:example_aggr_prop}(a)(b) shows how the inf-based and upd-based approaches work. The green and orange scope denotes the affected nodes of query and update requests, respectively. In a word, the two approaches perform either $L$-layer forward inference or $L$-layer backend update, which are both slow operations according to Section \ref{subsec:weakness}. Note that using existing update strategies~\cite{cb}, one more hop of neighbors are needed for $L$-layer backend update (i.e., $L$-layer update involves $L+1$-hop neighbors). See Section~\ref{sec:aip} for more details.

By CSM, the handlers for query and update requests collaborate to complete the entire serving process. The workloads allocated to the two handlers are balanced by the value of $M$.
Taking Figure \ref{fig:example_aggr_prop}(c) as an example, by applying collaborative serving ($M=1$ in the example), $L-M$(i.e., $2-1=1$) layers are offloaded to the query handler. In this case, the backend update involves $M+1$(i.e., 2)-hop neighbors, while the forward inference involves $L-M$(i.e., 1)-hop neighbors.

Table \ref{table:num_nodes_cb} shows the number of nodes involved in the computation. 
In the table, $c$ denotes the average connectivity of graph, $L$ denotes the number of layers of GNN model. As existing update strategies need one more hop of neighbors, it results in the $L+1$ or $M+1$ in exponent. 
$\star$ in Table~\ref{table:num_nodes_cb} indicates the special situation that if $M=0$, the value will be 1 rather than $c$. 
\begin{table}
\setlength{\belowcaptionskip}{5.pt}
\centering
\caption{The number of nodes involved in the computation of different approaches (i.e., the computational complexity).
}
\label{table:num_nodes_cb}
\footnotesize
\begin{tabular}{@{}lll@{}}
\toprule
           & \textbf{Query}     & \textbf{Update}        \\ \midrule
\textbf{Inf-based} & $c^L$     & 1             \\
\textbf{Upd-based} & 1         & $c^{L+1}$     \\
\textbf{STAG: CSM}   & $c^{L-M}$ & ${c^{M+1}}^{\star}$ \\ \bottomrule

\end{tabular}
\end{table}

In our design, we propose for the first time that the calculation of GNNs can be divided into two parts in $L$ layers, effectively balancing staleness and latency. 

\subsection{Coordinating The Inference and Update}

\subsubsection{Calculating the Inference and Update Time}
Let $t_q(M)$ and $t_u(M)$ represent the processing time of query and update requests with a given $M$, respectively. Obviously, $t_q(M)$ and $t_u(M)$ should be proportional to the number of nodes involved in the computation.
\begin{equation}
\label{eq:qtime_utime}
\begin{aligned}
& t_q(M) \propto c^{L-M} \\
& t_u(M) \propto c^{M+1} \ \mathbf{if} \ M \neq 0 \ \mathbf{else} \ 1
\end{aligned}
\end{equation}


According to Equation~\eqref{eq:qtime_utime}, we can observe that for a given $M$, $t_q(M)$ is a $(L-M)$-order polynomial function on $c$, and $t_u(M)$ is an $(M+1)$-order polynomial function on $c$ (if $M \neq 0$; otherwise, $t_u(M)$ will be a constant). This indicates that it is possible to profile $t_q(M)$ and $t_u(M)$ with polynomial fitting as Equation~\eqref{eq:profile}.
\begin{equation}
\label{eq:profile}
\begin{aligned}
& t_q = \mathrm{poly}^{L-M}(c), \\
& t_u = \mathrm{poly}^{M+1}(c) \ \mathbf{if} \ M \neq 0 \ \mathbf{else} \ \mathrm{poly}^{0}(c),
\end{aligned}
\end{equation}

\begin{figure}
\centerline{\includegraphics[width=.9\linewidth]{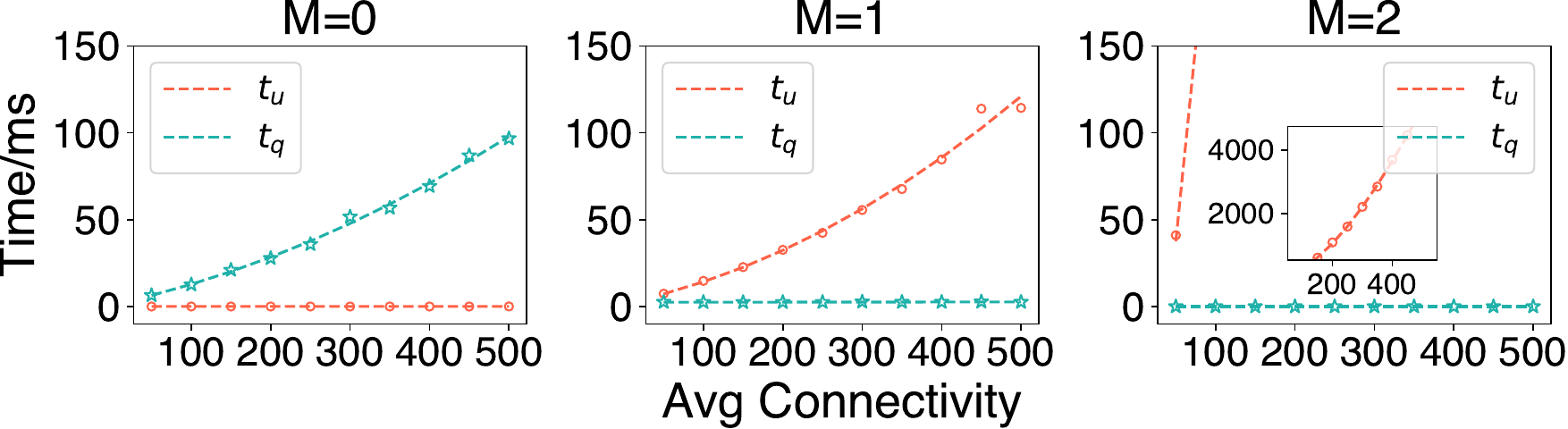}}
\caption{Profiled curves of $t_q(M)$ and $t_u(M)$ on Reddit dataset, with a 2-layer GCN model (i.e., $L=2$). Considering $t_u$, for $M=0$, it is constant (i.e., $0$-order function); for $M=1$, it is a $2$-order function; for $M=2$, it is a $3$-order function. They meet the formalization of Equation ~\eqref{eq:profile}.} 
\label{fig:profile_example}
\end{figure}

Note that we profile $t_q(M)$ and $t_u(M)$ on $c$ because the average connectivity of nodes keeps evolving in a real-world dynamic graph. Figure~\ref{fig:profile_example} shows an example of polynomial fitting, which proves the rationality of our formalization.

\subsubsection{Determining The Appropriate $M$}
\label{sec:tuning}
Even with $t_q$ and $t_u$, it is still nontrivial to determining the appropriate $M$ that minimizes the staleness time and the response latency.
From the perspective of performance efficiency, the latency and staleness have different importance related to their workloads.
For instance, if the query requests are much more than update requests, we shall pay more attention to the latency rather than staleness of the system, and we should choose a bigger $M$.

Suppose the workloads, measured in terms of Request-per-Second (RPS), of query requests and update requests are $\mathrm{rps}_q$ and $\mathrm{rps}_u$, respectively. 
We determining the appropriate $M$ by defining and solving the optimization problem in Equation~\eqref{eq:optimization}.
\begin{equation}
\label{eq:optimization}
\begin{aligned}
\mathrm{min:} \ \ & \mathrm{rps}_q \cdot t_q(M) + \mathrm{rps}_u \cdot t_u(M) \\
\mathrm{s.t.:} \ \ & M = 0, 1, \cdots, L.
\end{aligned}    
\end{equation}


To solve the optimization problem,the coordinator profiles $t_q(M)$ and $t_u(M)$ according to Equation~\eqref{eq:profile}. It only needs to profile them once because the profiled function remains invariant despite of the changes of the graph and workload.
Then, it keeps tracking the characteristics of workload and graph (i.e., $\mathrm{rps}_q$, $\mathrm{rps}_u$, and the average connectivity $c$).
Lastly, based the profiled function and tracked data, it solves the optimization problem in Equation~\eqref{eq:optimization} using Algorithm~\eqref{alg:opt} and comes up with the appropriate $M$ dynamically.

The experimental results in Section~\ref{subsec:perf_csm} show that the coordinator has succeeded in tuning the appropriate $M$ dynamically, despite the changes of graphs and workloads. Thanks to CSM, STAG jumps out the `staleness-latency' dilemma and supports 2.7x$\sim$27x workload, compared to the existing approaches.

\begin{algorithm}[t]
\caption{The Optimization Algorithm of CSM}
\label{alg:opt}
    \SetKwFunction{Sample}{Sample}
	\KwIn{The profiled function of $t_q(M)$, $t_u(M)$ in Equation~\eqref{eq:profile}; Average connectivity of current graph $c$; workload RPS $\mathrm{rps}_q$, $\mathrm{rps}_u$; Number of layers $L$.
	}
	\KwOut{Appropriate value of $M$ that minimize the object function in Equation~\eqref{eq:optimization}}  
	\BlankLine

	\For{$M=0$ to $L$}
    {
        Calculate the predicted $t_q(M)$ and $t_u(M)$ according to the profiled function in Equation~\eqref{eq:profile};
        
        Calculate the value of objective function $\mathrm{Obj}_M$ according to Equation~\eqref{eq:optimization};
    }
    
    \textbf{return} \ $M$ that minimize $\mathrm{Obj}_M$ 
\end{algorithm}

\section{Additivity-based Propagation}\label{sec:aip}
In this section, we focus on accelerating the update phase in CSM. With the consideration of the duplicated computation in the existing update strategy, we propose an Additivity-based Incremental Propagation (AIP) strategy, that utilizes the additivity property to eliminate the duplicated computation.

\begin{figure}
\centerline{\includegraphics[width=.7\linewidth]{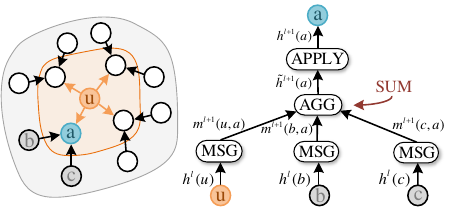}}
\caption{Illustration of propagation strategy. An update occurs at node $u$ influences the nodes in orange area. The representation of node $a$ is re-computed following the steps on the right hand side, involving uninfluenced nodes $b, c$.} 
\label{fig:cb}
\end{figure}

As mentioned in Section \ref{sec:cserving}, existing update strategy~\cite{cb} needs one more hop of neighbors for $L$-layer backend update. Because of the neighbor explosion problem, the one more hop of neighbors may lead to exponential increase of the computational complexity, severely slowing down the update.

Figure~\ref{fig:cb} illustrates how existing update strategy works. It caches the representations of nodes at each layer (denoted by $h^l$, $1 \le l \le L$). Thus, to re-compute $h^{l+1}$ of influenced nodes, it must follow the steps on the right hand side of Figure~\ref{fig:cb}, which requires $h^l$ of their 1-hop neighbors.
Obviously, the computation of nodes $b$ and $c$ is duplicated, since $h^l(b)$ and $h^l(c)$ are not changed. On a graph with high connectivity, it may cause hundreds of duplicated computation to re-compute one single node. 

\subsection{The Additivity Property}
By analyzing current GNN models, we observe an additivity property on the {\it AGG} phase of message passing paradigm in many typical GNN models (e.g., GCN~\cite{gcn}, GIN~\cite{gin}, SAGE~\cite{sage}, etc.). 

We leverage the property to eliminate the duplicated computation in existing strategy. 
We first illustrate the basic idea using typical GCN model, and generalize it later on. In GCN, the {\it AGG} phase is performed with SUM operator Equation~\eqref{eq:aggr_gcn}.
\begin{equation}
\label{eq:aggr_gcn}
  \widetilde{h}^{l+1}(v)=\sum_{u \in \mathrm{IN}(v)}m^{l+1}(u,v).
\end{equation}

Thanks to the additivity property of SUM operator,
when one of $m^{l+1}(u,v)$ changes (e.g., $m^{l+1}(u,a)$ in Figure \ref{fig:cb}) due to the graph dynamics happened at time $t+1$, we can re-compute $\widetilde{h}^{l+1}_{t+1}(v)$ according to the increment of $m^{l+1}(u,v)$ at time $t+1$. For instance, in Figure \ref{fig:cb}, the representation of node $a$ can be re-computed as in Equation~\eqref{eq:additivity_gcn}.
\begin{equation}
\label{eq:additivity_gcn}
\begin{aligned}
  \Delta m^{l+1}_{t+1}(u,a) & = m^{l+1}_{t+1}(u,a) - m^{l+1}_{t}(u,a) \\
  \widetilde{h}^{l+1}_{t+1}(a) & =\widetilde{h}^{l+1}_{t}(a) + \Delta m^{l+1}_{t+1}(u,a) 
\end{aligned}
\end{equation}

In the equation, $\Delta m^{l+1}_{t+1}(u,a)$ is the increment of $m^{l+1}(u,a)$ at time $t+1$. Note that we omit the MSG and APPLY phases in Equation~\eqref{eq:additivity_gcn}, which can be found in Equation~\eqref{eq:msg_passing}.
According to Equation~\eqref{eq:additivity_gcn}, we are able to re-compute the representation of influenced node $a$ without relying on the uninfluenced nodes $b, c$,  eliminating the unnecessarily duplicated computation in existing strategy.


\subsection{The Generalizability of AIP}
The strict application scope of AIP expressed using mathematical formulation is as follows: the {\it AGG} phase can be decomposed into binary operations that form an abelian group in the set of real numbers. In simple terms, this means that the operations in the {\it AGG} phase satisfy commutativity and additivity. SUM and MEAN are two typical {\it AGG} operations that satisfy this requirement.

\begin{table}[]
\setlength{\belowcaptionskip}{5.pt}
\caption{Extended Table \ref{table:num_nodes_cb}.
$\star$ indicates the special situation: if $M=0$, the value will be 1 rather than $c^{M+1}=c$.
An example with $c=100$, $L=2$, $M=1$ is presented for intuition.
}
\label{table:num_nodes_add}
\footnotesize
\setlength{\tabcolsep}{1mm}
\resizebox{\columnwidth}{!}{%
\begin{tabular}{@{}cccccc@{}}
\toprule
             & \multicolumn{2}{c}{\textbf{Query}} & \multicolumn{2}{c}{\textbf{Update}}                                                         & \textbf{Total}   \\ \midrule
             & theory       & example    & theory                                                                   & example & example \\ \midrule
\textbf{Inf-Based}   & $c^L$        & $10^4$           & 1                                                                        &  1       &  $10^4$       \\
\textbf{Upd-Based}   & 1            & 1           & $c^{L+1}$                                                                & $10^6$        & $10^6$        \\
\textbf{STAG: AIP}    & 1            & 1           & $c^L$                                                                    & $10^4$        & $10^4$        \\
\textbf{STAG: CSM}     & $c^{L-M}$    & $10^2$           & ${c^{M+1}}^*$ & $10^4$        & $1.01\times 10^4$        \\
\textbf{STAG: CSM+AIP} & $c^{L-M}$    & $10^2$            & $c^M$                                                                      & $10^2$         & $2 \times 10^2$         \\ \bottomrule
\end{tabular}
}
\end{table}

Although there are some operations that do not intuitively satisfy this property, they can be applied by AIP with simple modifications. For MIN and MAX operations, the extreme value (minima/maxima) can be cached as $\widetilde{h}$. In most update requests, we can maintain $\widetilde{h}$ incrementally. Only in rare case of deleting the edge corresponding to the extreme value, it degenerates to a full aggregation. Furthermore, for GATs~\cite{gat} that utilizes the attention mechanism in {\it AGG} operations, the denominator of the softmax operation can be extracted, and then the numerator will also exhibit additivity.

According to previous studies {~\cite{implications}}, the {\it AGG} operation of sum, mean, max types can reach a percentage of $83.02\%$ in the message passing paradigm, while the percentage of {\it AGG} operaton of the attention mechanism is $13.2\%$. For nonlinear operations that cannot be transformed, AIP degenerates to the naive update mechanism. In this case, the CSM mechanism is still able to take effect though AIP cannot provide an acceleration on the update phase.

\subsection{Applying AIP on Collaborative Serving}
\label{subsec: apply_aip}
Based on the newly identified additivity property,
we apply AIP to speed up the update phase.
To realize AIP, graph maintainer in Figure~\ref{arch} maintains intermediate results of {\it AGG} phase (i.e., $ \widetilde{h}^{l}(v)$) in the cache manager. Then, the system can perform update incrementally according to Equation~\eqref{eq:additivity_gcn}, without involving any unaffected node.

Table~\ref{table:num_nodes_add} illustrates the effect of AIP in reducing computation overhead. 
As observed, the computation overhead decreases exponentially, by applying AIP strategy in the update phase. 
This is because AIP eliminates the update's dependency on unaffected nodes.

\section{Evaluation of STAG}
\label{sec:evaluation}


\subsection{Experimental Setup and Baselines}



The experimental setup of STAG is shown in Table \ref{tab:setup}. The experiments are performed on a Xeon(R) Silver 4210 CPU and an NVIDIA A100 GPU. 
Table~\ref{tab:datasets} shows the statistics of the six used graphs. 
We use the two dynamic datasets to simulate the dynamic nature of graphs in real world. While using static datasets is mainly because their size is much larger than publicly available dynamic datasets so as to demonstrate the scalability of STAG. 


We compared STAG with Aligraph~\cite{aligraph} and CB~\cite{cb} that use inf-based approach and upd-based approach, respectively. As CB mainly focuses on the cache selection strategies, we implement CB by applying its representation update method in STAG.


\begin{table}[]
\renewcommand{\arraystretch}{2}
\setlength{\belowcaptionskip}{5.pt}
\setlength{\tabcolsep}{2mm}
\caption{The hardware, software, and benchmarks.}
\label{tab:setup}
\resizebox{\columnwidth}{!}{%
\Huge
\begin{tabular}{c|ccc}
\hline
\textbf{Hardware}     & \multicolumn{3}{c}{CPU: Xeon(R) Silver 4210R; GPU: NVIDIA A100; DRAM: 256GB}         \\ \hline
\textbf{OS \& Driver} & \multicolumn{3}{c}{Ubuntu: 20.04.1 (kernel 5.15.0)}                                       \\ \hline
\textbf{Software}     & \multicolumn{3}{c}{Python 3.8.13; Pytorch 1.11.0; DGL 0.8.1; CUDA: 11.3}                              \\ \hline
\textbf{Models}       & \multicolumn{3}{c}{GCN~\cite{gcn}; GIN~\cite{gin}; GraphSAGE~\cite{sage}}                                                   \\ \hline
\textbf{Datasets}     & \multicolumn{1}{c|}{Reddit~\cite{reddit}; Products~\cite{ogb}}      & \multicolumn{1}{c|}{Citation2~\cite{ogb}; PPA~\cite{ogb}}    & \multicolumn{1}{c}{dReddit~\cite{jodie}; dWiki~\cite{jodie}}  \\ \hline 
\textbf{Task type}    & \multicolumn{1}{c|}{Node Classification(NC)}            & \multicolumn{1}{c|}{Link Prediction(LP)}  & \multicolumn{1}{c}{NC \& LP} \\ \hline
\end{tabular}%
}
\end{table}

\begin{table}
\centering
\caption{Dataset statistics.}
\label{tab:datasets}
\footnotesize
\begin{tabular}{@{}ccccc@{}}
\toprule
         & \textbf{Nodes} & \textbf{Edges}  & \textbf{Connectivity} & \textbf{Feat. Dim.} \\ \midrule
\textbf{Reddit}   & 233.0K & 114.6M & 492.0 & 602       \\
\textbf{Products} & 2.4M   & 123.7M & 50.5 & 100        \\
\textbf{Citation2} & 2.9M   & 60.7M  & 20.7 & 128         \\
\textbf{PPA}      & 576.3K & 42.5M  & 73.7 & 58        \\
\textbf{dReddit}      & 11K & 1.3M  & 122.44 & 172        \\
\textbf{dWiki}      & 9K & 315K  & 34.13 & 172        \\ \bottomrule
\end{tabular}
\end{table}

\subsection{Implementation Details}
\label{subsec:implementation}

We implement STAG based on DGL~\cite{dgl}, because it provides extensive GNN-related interface. 
To implement CSM, we divide an $L$ layers GNN model into an $M$-layer update sub-model and an $L-M$ layer query sub-model that can be invoked independently.
To implement AIP, we customize the message passing scheme for the $M$-layer update part of each GNN model. Different from traditional message aggregation among all neighbors~\cite{cb}, AIP only propagates delta of messages, thus greatly reducing the computation. 

Considering large-scale graph in production cannot fit in GPU, we store the original graph in CPU memory. The operations on the original graph such as sampling and caching are performed on CPU as well. On the other hand, the sampled subgraphs are relatively small so that we utilize GPU to accelerate the model inference on them.

\subsection{The Staleness and The Latency}
\label{subsec:perf_STAG}


\begin{figure}
\centerline{\includegraphics[width=.95\linewidth]{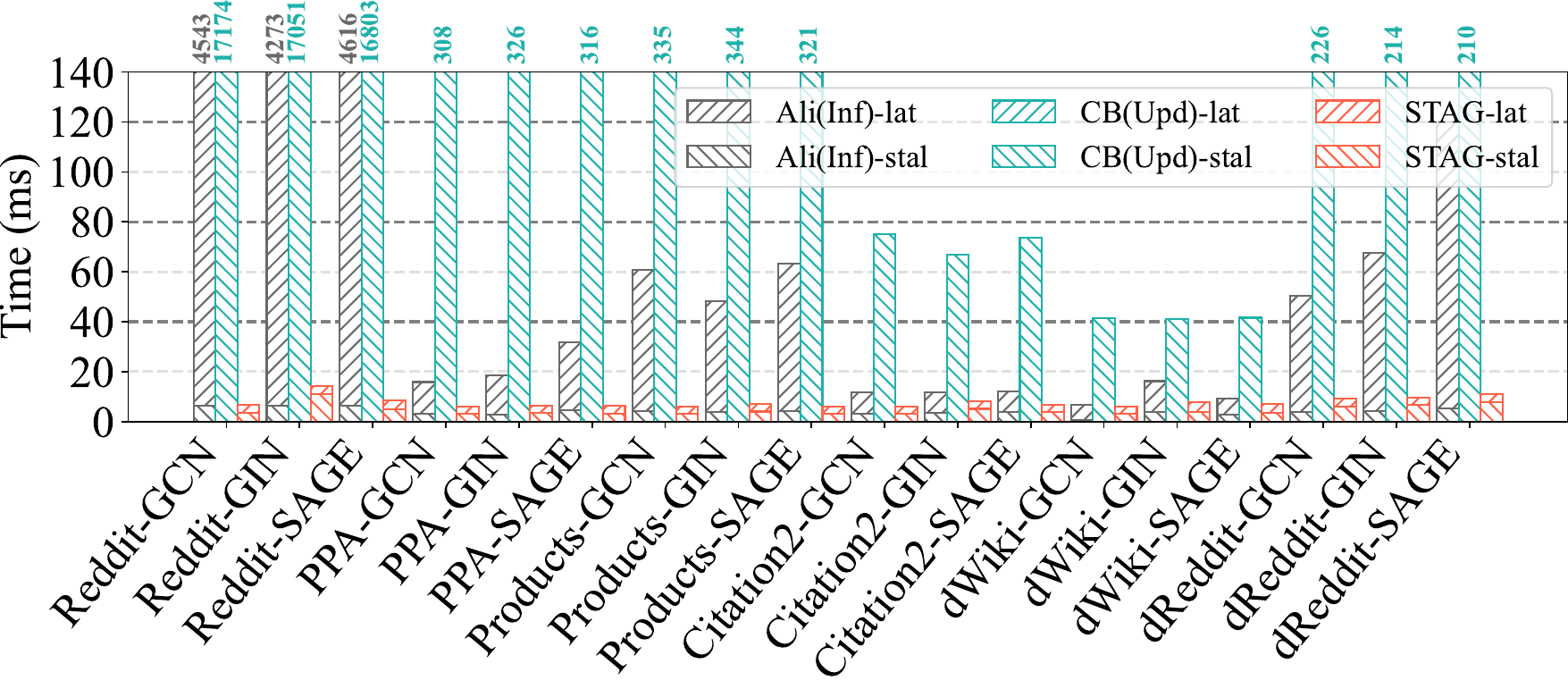}}
\caption{The latency and the staleness time of 18 test cases with inf-based approach, upd-based approach, and STAG.
}
\label{fig:STAG_lat_stal_12cases}
\end{figure}

Figure~\ref{fig:STAG_lat_stal_12cases} shows the response latency and the staleness time of all the $3\times 6$ test cases (3 GNN model choices and 6 graph choices). The load of update and query requests are both 150 RPS. Compared with CB (upd-based approach), STAG greatly reduces the staleness time by 37$\sim$17171 milliseconds, while only increasing the response latency to 3$ \sim $4 milliseconds. 

Obviously, STAG significantly reduces the staleness time, with only slight increase in the response latency.
STAG achieves both the low staleness and the low latency, because it reduces the computation overhead of graph serving through collaborative serving, and speeds up the update process.
The result is also consistent with the theoretical analysis that the computation overhead of STAG has been exponentially reduced compared with the two approaches.

\subsection{Supported Peak Workload}

\begin{figure}
\centerline{\includegraphics[width=.45\linewidth]{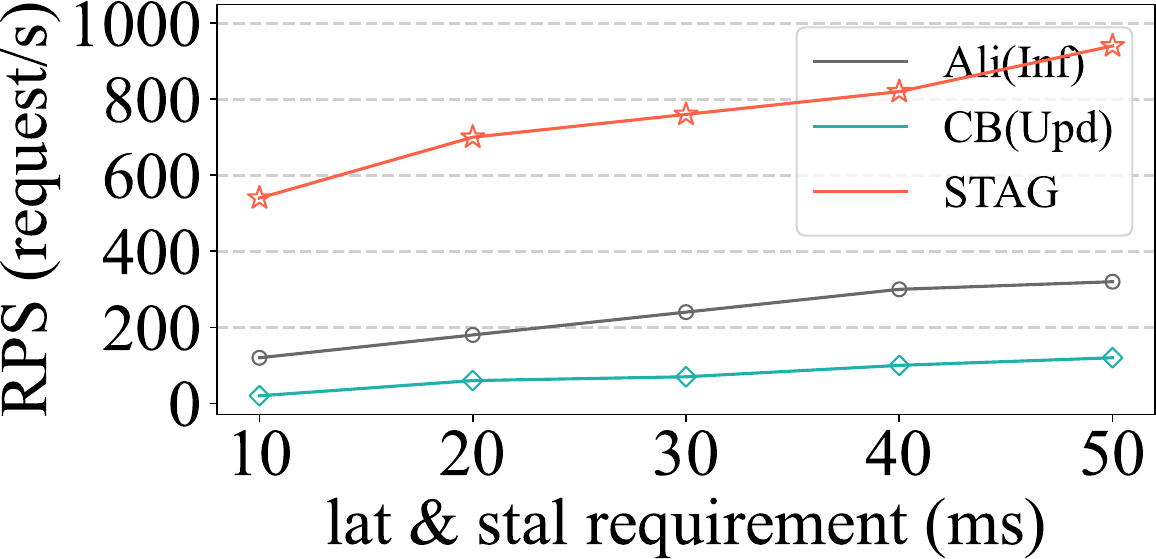}}
\caption{The peak supported RPS with different staleness and latency requirements. 
} 
\label{fig:STAG_rps_sup}
\end{figure}

Figure \ref{fig:STAG_rps_sup} shows the supported peak load with given latency and staleness time requirements. 
In the figure, the $x$-axis shows the response latency and staleness requirements, and the $y$-axis shows the supported peak loads with inf-based approach, upd-based approach, and STAG.
As observed, STAG supports 2.7$\sim$4.5x and 7.8$\sim$27x higher loads than Aligraph (inf-based system) and CB (upd-based system), respectively.

\subsection{Performance of Collaborative Serving}
\label{subsec:perf_csm}

In this subsection, we verify the adaptability of STAG in two aspects: 1. {\it Different Load Patterns}. 2. {\it Graph Evolution}.
With {\bf CSM}, STAG should dynamically tune an appropriate $M$ by solving the optimization problem in Equation \eqref{eq:optimization}.

\begin{figure}
    \centering
    \subfigure[Load Patterns]
    {
    \includegraphics[width=0.48\linewidth]{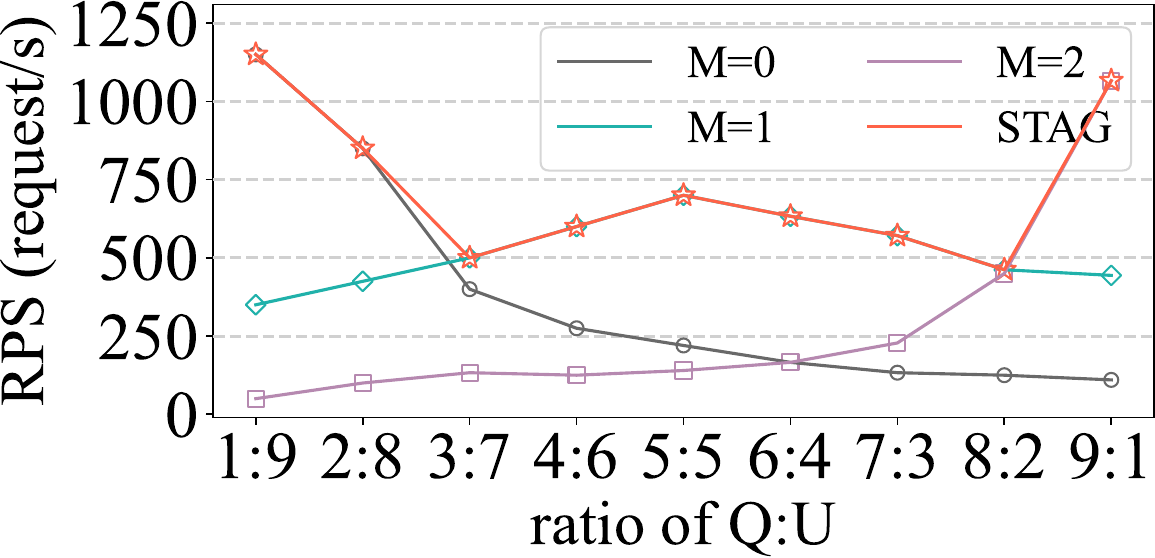}
    }
    \subfigure[Graph Evolution]{
    \includegraphics[width=0.44\linewidth]{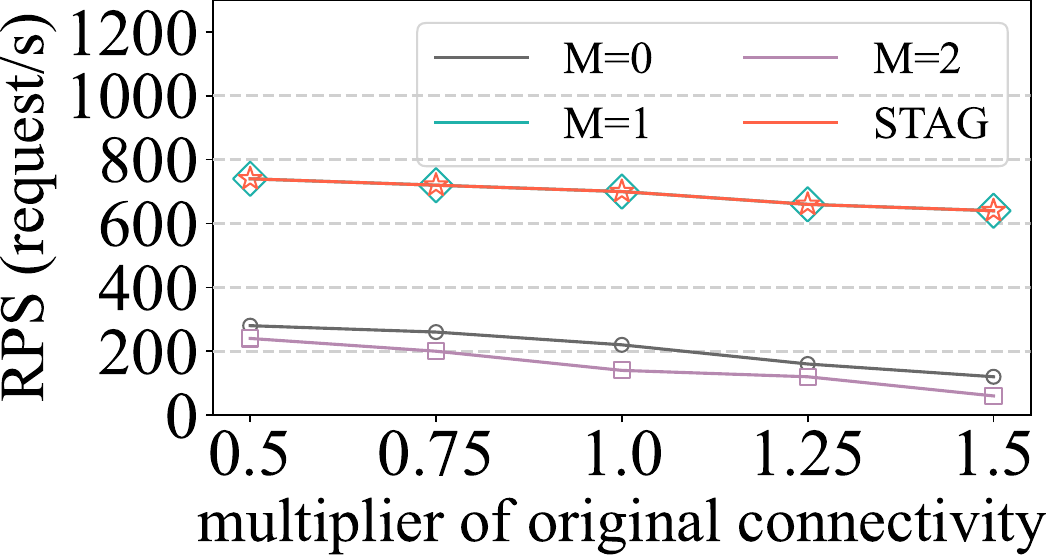}
    }
    \caption{The supported peak RPS under different circumstance.}
    \label{fig:STAG_rps_ratio_var_sup}
\end{figure}

Figure \ref{fig:STAG_rps_ratio_var_sup} shows the experimental results under test case GCN + Products. In Figure \ref{fig:STAG_rps_ratio_var_sup}(b), we randomly add/delete edges to simulate the evolution of real-world graph.
As observed, STAG can always find the best $M$ to balance the load of serving handler and graph maintainer, supporting higher workloads (RPS).
Experiments under other test cases also show similar phenomenon. 
This proves that the collaborative serving mechanism in STAG is adaptive.

\subsection{Performance of Incremental Propagation}
\label{subsec:perf_aip}
In this subsection, we compare our additivity-based incremental propagation (AIP) with the update strategy in CB~\cite{cb}.

\begin{figure}
\centerline{\includegraphics[width=.8\linewidth]{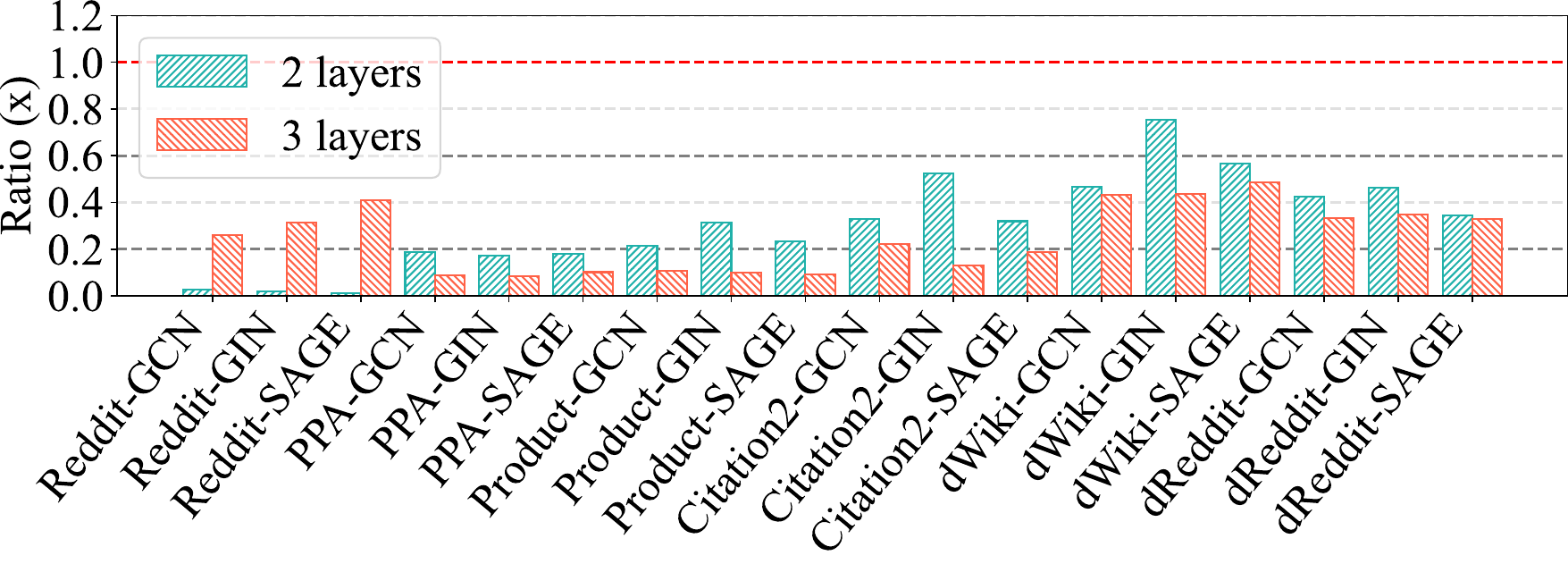}}
\caption{The processing time of the update phase with AIP, normalized to the time with CB. 
} 
\label{fig:aip_time}
\end{figure}


Figure~\ref{fig:aip_time} shows the processing time of the update phase with AIP, normalized to the time with CB.
As observed, AIP accelerates the update phase by 1.3x$ \sim $90.1x.
The speedup ratio is calculated 
by $\mathrm{ratio}_{sp} = {t_\mathrm{baseline}} / {t_\mathrm{AIP}}$.

As observed from Figure \ref{fig:aip_time}, 
there are two general trends of $\mathrm{ratio}_{sp}$.
First of all, higher graph connectivity leads to higher $\mathrm{ratio}_{sp}$ (see Table~\ref{tab:datasets} for the connectivity of datasets).
Secondly, $\mathrm{ratio}_{sp}$ tends to be higher in deeper models (i.e., more layers). 
Note that Reddit is an exception, where $\mathrm{ratio}_{sp}$ for 3 layers is not higher than that for 2 layers. This is because Reddit has a very high connectivity while a relatively small graph size (see Table~\ref{tab:datasets}), so 2-hop neighbors already cover most of the nodes, and expanding to 3-hop neighbors does not cause an exponential increase in node size. This situation is unlikely to happen on real-world graphs, which are typically large and sparse.

\subsection{Scalability}
\label{subsec:scale}

\begin{figure}
    \centering
    \subfigure[latency and staleness]
    {
    \includegraphics[width=0.41\linewidth]{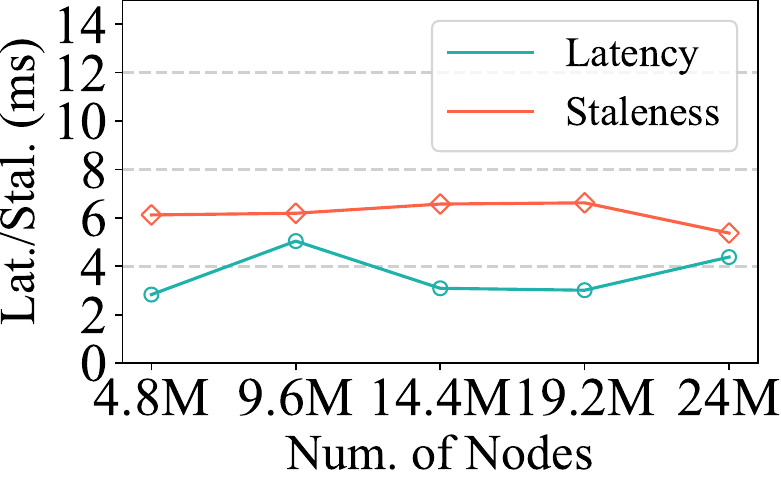}
    }
    \hspace{3mm}
    \subfigure[memory consumption]{
    \includegraphics[width=0.47\linewidth]{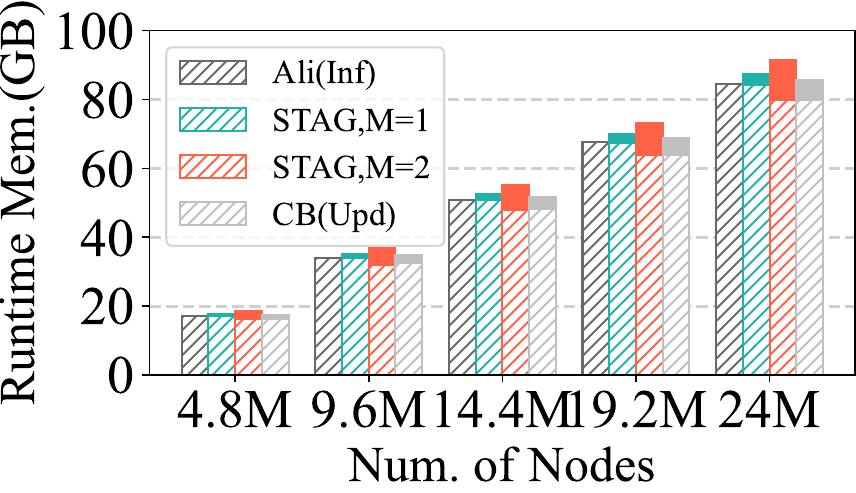}
    }
    \caption{The changes in latency/staleness and memory consumption as the graph size increases.
    }
    \label{fig:scal}
\end{figure}

To verify the scalability of STAG, 
We artificially increased the size of Products~\cite{ogb} by adding nodes and edges while keeping its connectivity constant. Figure~\ref{fig:scal} shows that there is no significant change in staleness and latency as the graph size increases, while memory consumption increases linearly with the graph size. This indicates that STAG is scalable as long as there is enough memory. When the graph size becomes too large to be stored in a single machine's memory, it can be stored on multiple devices in a distributed manner using techniques such as graph partitioning, which is orthogonal to STAG.

\subsection{Ablation Study}

\begin{figure}
\centerline{\includegraphics[width=.95\linewidth]{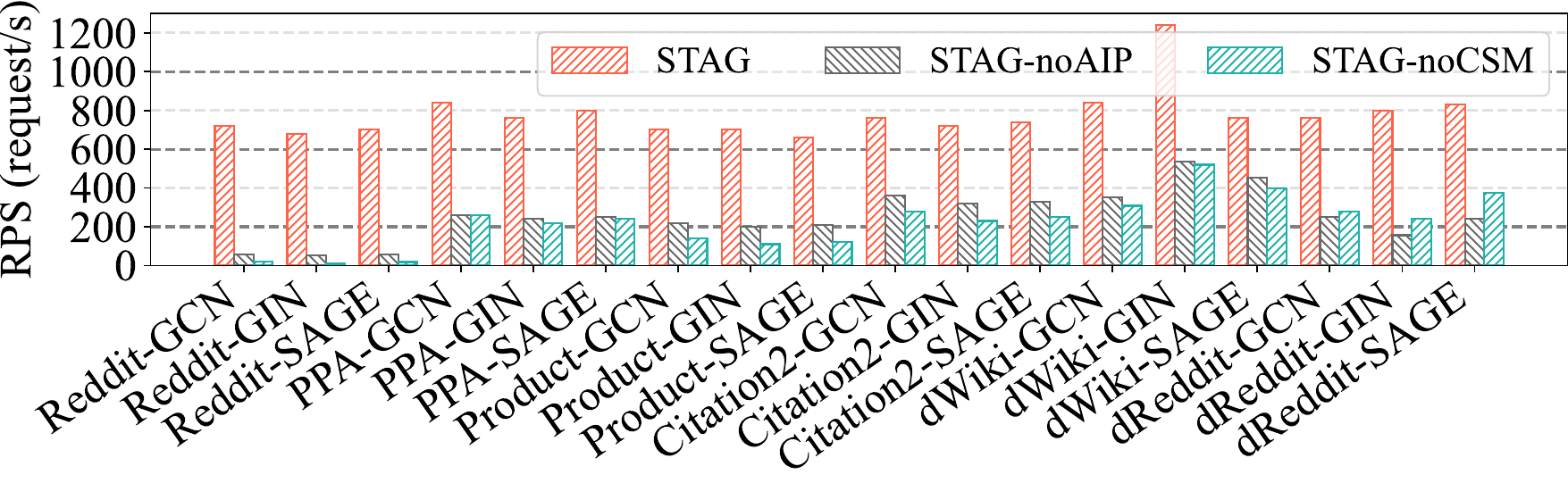}}
\caption{Ablation study of CSM and AIP. 
} 
\label{fig:ablation_rps_sup}
\end{figure}

Figure~\ref{fig:ablation_rps_sup} shows the supported peak loads with 18 test cases. 
In the figure, STAG-noAIP and STAG-noCSM are variants of STAG that disable the incremental propagation strategy and collaborative serving mechanism, respectively. As observed, the supported peak load decreases significantly if any component in STAG is disabled.
Both the two techniques contribute to the good performance of STAG.

\section{Conclusions}

In this work, we propose a dynamic GNN serving framework {\bf STAG} that
achieves low staleness and low latency through a {\it collaborative serving mechanism} ({\bf CSM}) and an {\it additivity-based incremental propagation strategy} ({\bf AIP}).
With {\bf CSM}, only part of the node representations are updated during the update phase, the final representations are calculated in the inference phase.
Moreover, {\bf AIP} reuses intermediate data during backend update, thus greatly accelerating the update phase.
STAG accelerates update phase by 1.3x$ \sim $90.1x, greatly reduces staleness with a slight increase in response latency.

\bibliographystyle{IEEEtran}
\bibliography{reference}

\end{document}